\documentclass[10pt,conf]{IEEEtran}
\usepackage{geometry}
\usepackage{bbold}
\usepackage{subcaption}%
\usepackage[cmex10]{amsmath}
\usepackage{amssymb}
\usepackage{verbatim}
\usepackage{array}
\usepackage{latexsym}
\usepackage{color}
\usepackage{textgreek}
\usepackage{subscript}
\usepackage{rotating}
\usepackage{booktabs,multirow}
\usepackage{mdframed}	
\usepackage{tabularx}
\usepackage{amsmath}
\usepackage{makecell}
\usepackage{tabto}

\usepackage{mathrsfs}
\usepackage{url}
\usepackage{cite}
\usepackage{listings}
\usepackage{array}
\usepackage[table]{xcolor}

\newcommand{\nref}[1]{\textcolor{red}{[{\bf NEED CITATION}]}}
\begin{document}
\setlength{\columnsep}{0.2 in}
\def\BibTeX{{\rm B\kern-.05em{\sc i\kern-.025em b}\kern-.08em T\kern-.1667em\lower.7ex\hbox{E}\kern-.125emX}}

\title{Collaborative Self Organizing Map with DeepNNs for Fake Task Prevention in Mobile Crowdsensing}

\author{
	Murat~Simsek,~\IEEEmembership{Senior Member,~IEEE},~Burak Kantarci,~\IEEEmembership{Senior Member,~IEEE},~Azzedine Boukerche,~\IEEEmembership{Fellow,~IEEE}
	\thanks{
	The authors are with the School of Electrical Engineering and 
        Computer Science at the University of Ottawa, Ottawa, ON, K1N 6N5, Canada.
		E-mail: \{ murat.simsek,burak.kantarci\}@uottawa.ca and {boukerch}@ site.uottawa.ca }

}

\maketitle
\thispagestyle{empty}
\pagestyle{empty}

\begin{abstract}
Mobile Crowdsensing (MCS) is a sensing paradigm that has transformed the way that various service providers collect, process, and analyze data. MCS offers novel processes where data is sensed and shared through mobile devices of the users to support various applications and services for cutting-edge technologies. However, various threats, such as data poisoning, clogging task attacks and fake sensing tasks adversely affect the performance of MCS systems, especially their sensing, and computational capacities. Since fake sensing task submissions aim at the successful completion of the legitimate tasks and mobile device resources, they also drain MCS platform resources. In this work, Self Organizing Feature Map (SOFM), an artificial neural network that is trained in an unsupervised manner, is utilized to pre-cluster the legitimate data in the dataset, thus fake tasks can be detected more effectively through less imbalanced data where legitimate/fake tasks ratio is lower in the new dataset. After pre-clustered legitimate tasks are separated from the original dataset, the remaining dataset is used to train a Deep Neural Network (DeepNN) to reach the ultimate performance goal. Pre-clustered legitimate tasks are appended to the positive prediction outputs of DeepNN to boost the performance of the proposed technique, which we refer to as pre-clustered DeepNN (PrecDeepNN). The results prove that the initial average accuracy to discriminate the legitimate and fake tasks obtained from DeepNN with the selected set of features can be improved up to an average accuracy of 0.9812 obtained from the proposed machine learning technique.
\end{abstract}

\begin{IEEEkeywords}
\setlength{\columnsep}{0.2 in}
	Mobile Crowdsensing, Machine Learning, Self Organizing Feature Map, Deep Neural Networks, Fake Task Prevention, Sensing as a Service, \end{IEEEkeywords}

\IEEEpeerreviewmaketitle

\section{Introduction}
\label{sec:intro}Mobile crowdsensing (MCS) has been integrating a useful data collection capability into the smart environments by using embedded sensing capabilities of smart devices in either opportunistic or participatory manner \cite{Capponi.2019}. Smart devices equipped with various sensors are the building blocks of large scale sensing networks \cite{Yuanyuan.2021} and  mobility and sociability directly affect  the context of MCS data \cite{s21196397}. MCS use cases are various such as healthcare \cite{Guo-Safety.2017,Liu.2019}, public safety \cite{ELKHATIB2020222}, transportation, traffic, and environmental monitoring \cite{Huang.2019}.

MCS platforms exploit the ubiquity of smart mobile devices and assume that their sensing, processing, and communication resources are readily available anytime and anywhere. Interrupting the resource allocation adversary affects the task compilation and participation ratios by keeping user and platform utilities as high as possible during MCS campaign. Malicious tasks can be designed in an intelligent way to consume the resources of the devices as well as MCS servers in the platform \cite{Zhang2021Empowering}. There have been prior solutions that aims to prevent the malicious users from deceiving the MCS platform using game theoretic strategies \cite{Xiao.2018}. 
The literature lacks thorough studies on fake tasks, hence a detailed investigation is required. However, there are few useful studies, for instance, the authors in \cite{Simsek2021Utility} detect the fake tasks that target the attack surface of the MCS servers or devices, through machine learning algorithms.  In the event of a successful fake task submission attempt, fake tasks can drain the resources of both participating devices and MCS servers. Recent studies \cite{Simsek2021Utility, Simsek2020Detecting, Zhang2020Ensemble} have pointed out that ensemble techniques, if succeed in detecting and eliminating the fake tasks from the MCS platform, can achieve remarkable battery savings against such attacks, and lead to significant savings in battery lifetimes of the smart devices.

Previous works empowered a knowledge-based prior knowledge input method that employed a deep neural network (DeepNN) with four hidden layers in combination with a  Self Organizing Feature Map (SOFM) \cite{Zhang2021Empowering, Locally2020Chen} in order to enhance legitimacy detection performance in \cite{Simsek2020Self}. Moreover, the DeepNN performance had been shown to improve when combined with ReliefF-based sequential feature selection \cite{Sood2019Deep}. The motivation in this study is to cope with the class imbalance between legitimate and illegitimate tasks through early detection and elimination of legitimate tasks by empowering an unsupervised method in the training  dataset. Thus,  the  test  performance of the legitimacy detection model can be further improved. In the proposed technique, SOFM is used of pre-clustering legitimacy tasks  for both training and test datasets. As remaining mixed (legitimate and fake) tasks in the training and test datasets can be used for training and test of DeepNN model, it is referred to as pre-clustered DeepNN (PrecDeepNN).  After the test performance of  PrecDeepNN is completed, pre-clustered legitimate-only (PrecL) tasks in the test dataset are combined with test performance of PrecDeepNN to obtain the PrecDeepNNPrecL model over all results. Although SOFM is utilized in MCS applications such as intelligent attack anticipation \cite{Zhang2021Empowering, Locally2020Chen, Quintal2019Sensory}, this is the first time  SOFM driven pre-clustering of legitimate tasks is proposed to boost the overall performance of the DeepNN.

The paper is organized in six sections. Section II presents the motivation and the recent related studies. Section III introduces the characteristics of submitted tasks and their generation process. Section IV presents the proposed methodology in detail. Section V summarizes the simulation results and verifies the performance of the proposed methodology. Finally, concluding remarks and future work are presented in Section VI.

\begin{figure}[!hbt]
        \centering
        \includegraphics[width = 0.5\textwidth]{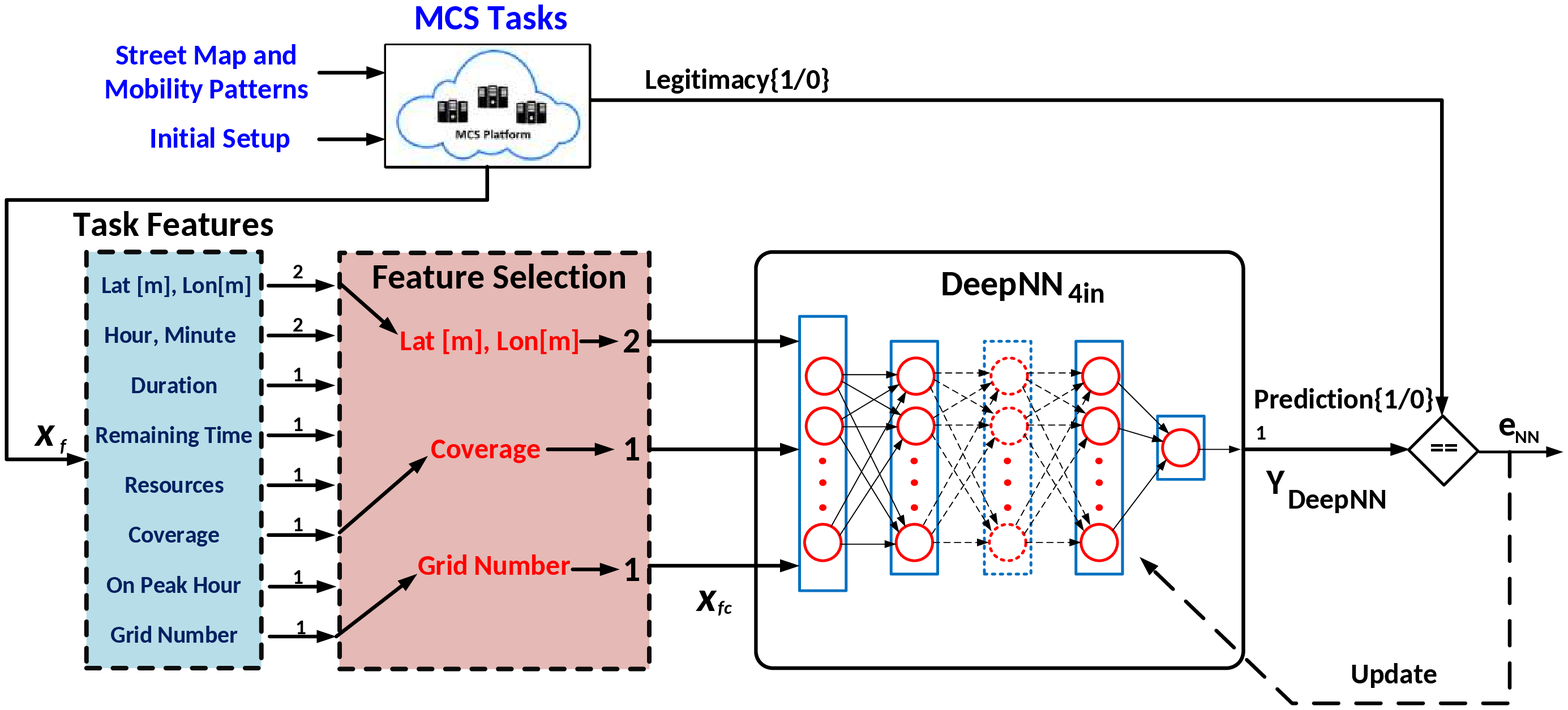}
        \caption{DeepNN training process for getting the initial accuracy of the performance evaluation using 4 selected features out of available 10 features provided by MCS tasks. }
        \label{fig:deep_nn}
\end{figure}

\section{Related Work and Motivation}
\label{sec:bg}

Despite their efficiency, MCS systems are vulnerable due to relying on the uncertainty of the availability of the distributed sensing devices and their heterogeneity \cite{Khan.2019} and potential sensing inaccuracy \cite{Dong.2021}.

The study in \cite{Miao.2018} studies the performance of machine learning classifiers against data poisoning attacks. Additionally, the authors of \cite{Gisdakis.2016} address intelligent malicious attacks with ensemble learning methods via context recognition through machine learning techniques. There are also recent studies that investigate identification and filtering of fake MCS tasks applying machine learning solutions  \cite{Simsek2021Utility, Chen2021Federated}. To ensure the security of MCS platforms, deep learning methods have also been studied \cite{Xiao.DL.2018, Sood2019Deep}. 

The contribution of this paper is transformative compared to the previous studies that were incremental improvements of illegitimate task detection: In this study, by applying SOFM-based clustering for early detection of legitimate tasks, the number of legitimate tasks is reduced in the training dataset. By doing so, the class imbalance of the training dataset is mitigated so to improve the training performance with the ultimate goal of  better performance on the test set. Once the SOFM-based clustering is applied to the test dataset by aiming at detecting legitimate only tasks (i.e., true positives) before pursuing prediction through the trained model.
After applying all machine learning strategies, the overall performance of the proposed methodology outperforms the baseline performance.

\begin{figure*}[!hbt]
        \centering
        \includegraphics[width = 0.58\textwidth]{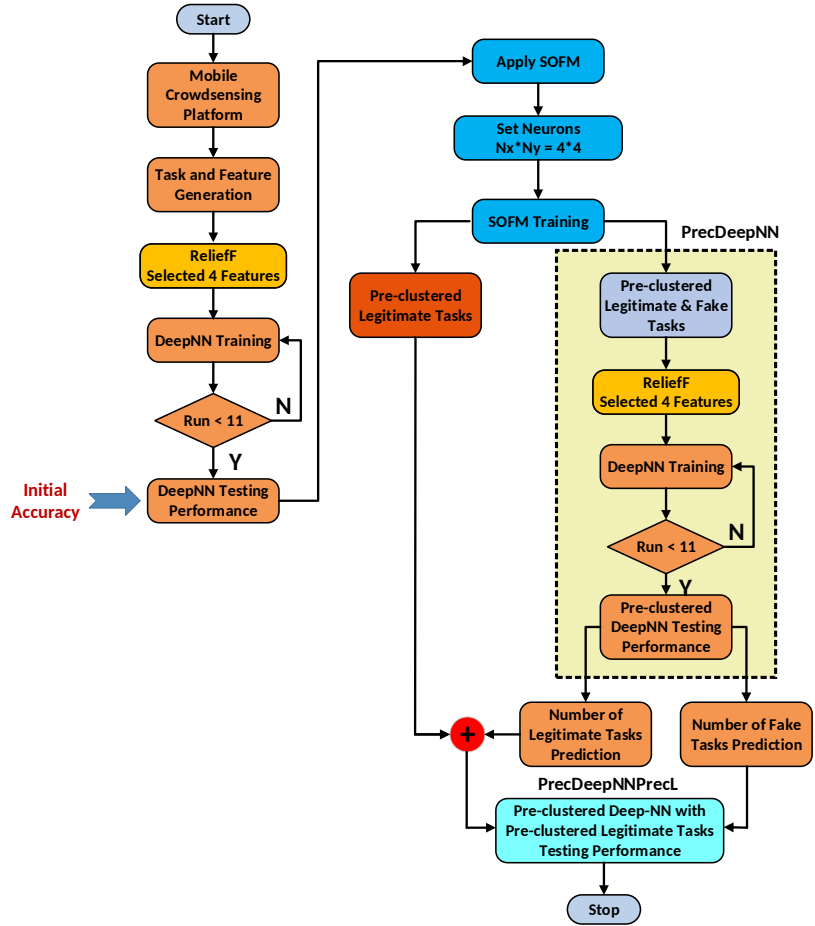}
        \caption{General algorithm for proposed methodology. }
        \label{fig:algorithm}
\end{figure*}

\section{Legitimate and Fake Crowdsensing Tasks Generation in MCS Platform }
Fake tasks are injected by adversaries to maliciously affect the performance of MCS campaigns. Thus, not only participants experience as dissatisfied but also MCS resources, such as smartphone sensors and computational capacity of MCS platform, are drained by adversarial attacks. To create defence mechanism against fake tasks, Crowdsensim simulator \cite{CrowdSensim} is used to simulate a realistic environment. Using this simulator, fake tasks should be designed to demonstrate the real impact on the MCS system. To achieve this purpose, fake tasks have been modelled by taking into account certain features as listed in Table \ref{tab:simulation} in \cite{Chen2021Federated}. 

\begin{table}[!hbt]
\centering
\caption{Simulation settings for task generation}
\label{tab:simulation}
\begin{tabular}{p{2.5cm}|p{2.3cm}|p{2.3cm}}
    \hline
    {}& Fake Tasks&Legitimate Tasks\\
    \hline
    Day&\multicolumn{2}{c}{Uniform  [1-6]}\\
    \hline
    Hour (task)&80\% : 7am to 11am;
    20\%: 12pm to 5pm& 8\%: 0pm to 5am; 92\%: 6am to 23pm\\
    \hline
  
    Duration (task)&70\% in \{40, 50, 60\} [min.]; 
    30\% in \{10, 20, 30\} [min.]&Uniform  \{10, 20, 30, 40, 50, 60\} mins\\
    \hline
    Battery consumption of smartphone &80\% in \{7\%-10\%\};
    20\% in \{1\% -6\%\}& Uniform \{1\%-10\%\}\\
    \hline
    \end{tabular}
\end{table}

All possible features for submitted tasks in MCS can be seen in Fig. \ref{fig:deep_nn}. 
The location of each task is defined as GPS coordinates such as 'latitude' and 'longitude'. Time-based features such as  'day', 'hour' and 'minute', determine the task-specific time requirements. The 'day' feature indicates the date among 6 days of the test week. Since this feature is varies from 1 to 6, and not specify distinctive property of legitimate and fake tasks, it is not used as one of the task's features. 'Duration' specifies the activation time of each task in minutes. "Resources"  defines the required battery level which is the percentage of consumption value of a smartphone by the tasks.'Legitimacy' is a binary label to identify legitimate tasks from fake tasks by using 1 for legitimate and 0 for fake tasks. 'Grid Number' defines grid based location of tasks that are covered by pre-defined square area. Moreover, to show the impact of busy communication time between  7am and 11am, "on-peak hours" is added as binary value referring whether the tasks is appeared in this time or not. To sum up all feature set, fake tasks are produced through statistical distribution  in Table \ref{tab:simulation}.

\section{Self Organizing Feature Map (SOFM) Driven pre-clustered Deep Neural Network (PrecDeepNN) Approach}
Machine learning has been mostly used as a multi-disciplinary technique to design intelligent strategy embedding the engineering design problems. These designs can be seen as intelligent attack or defence strategies in MCS systems. In this paper, machine learning methodology has been applied to detect legitimacy of MCS tasks which try to affect the performance of legitimate tasks during MCS campaigns. To obtain baseline classification performance as explained in the previous study \cite{Simsek2020Self}, ReliefF feature selection method is firstly applied to 10 meaningful features of the MCS dataset as depicted in Fig. \ref{fig:deep_nn}, then a Deep Neural Network (DeepNN) with 4 hidden layers outputs the average accuracy result of ten runs which are required to eliminate the random initial weights of DeepNN training process. 

SOFM was used for prior knowledge to reduce the complexity of DeepNN. In this study, SOFM emerges clustering performance to partition the legitimate-only tasks and mixed tasks with legitimate and fake. General work flow of the proposed methodology can be seen in Fig. \ref{fig:algorithm} regarding the algorithmic details. 
Each DeepNN structure is trained 10 times with random initial conditions
whereas the second part of the flow chart provides details about how to apply an unsupervised method to detect legitimate tasks to mitigate imbalance of legitimate and fake tasks. Further details are given in the following subsections.  

\begin{figure}[!hbt]
 \centering
    \begin{subfigure}{0.5\textwidth}
        \centering
        \includegraphics[width = 1\textwidth, trim=1cm 0.0cm 1cm 0.0cm,clip]{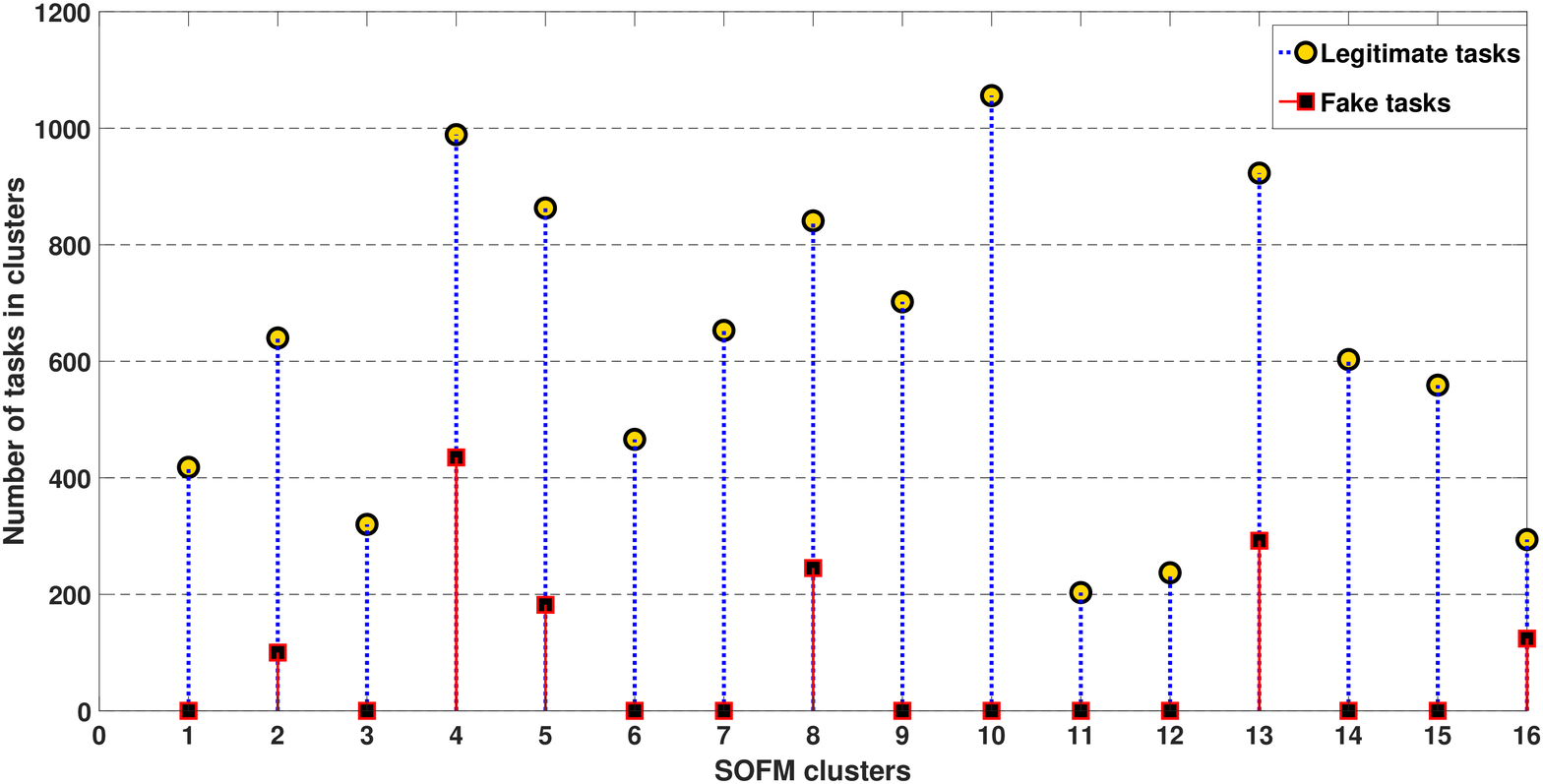}
        \caption{Training}
        \label{fig:sofm_clusters_train }
    \end{subfigure}
    \begin{subfigure}{0.5\textwidth}
        \centering
        \includegraphics[width = 1\textwidth, trim=1cm 0.0cm 1cm 0.0cm, clip]{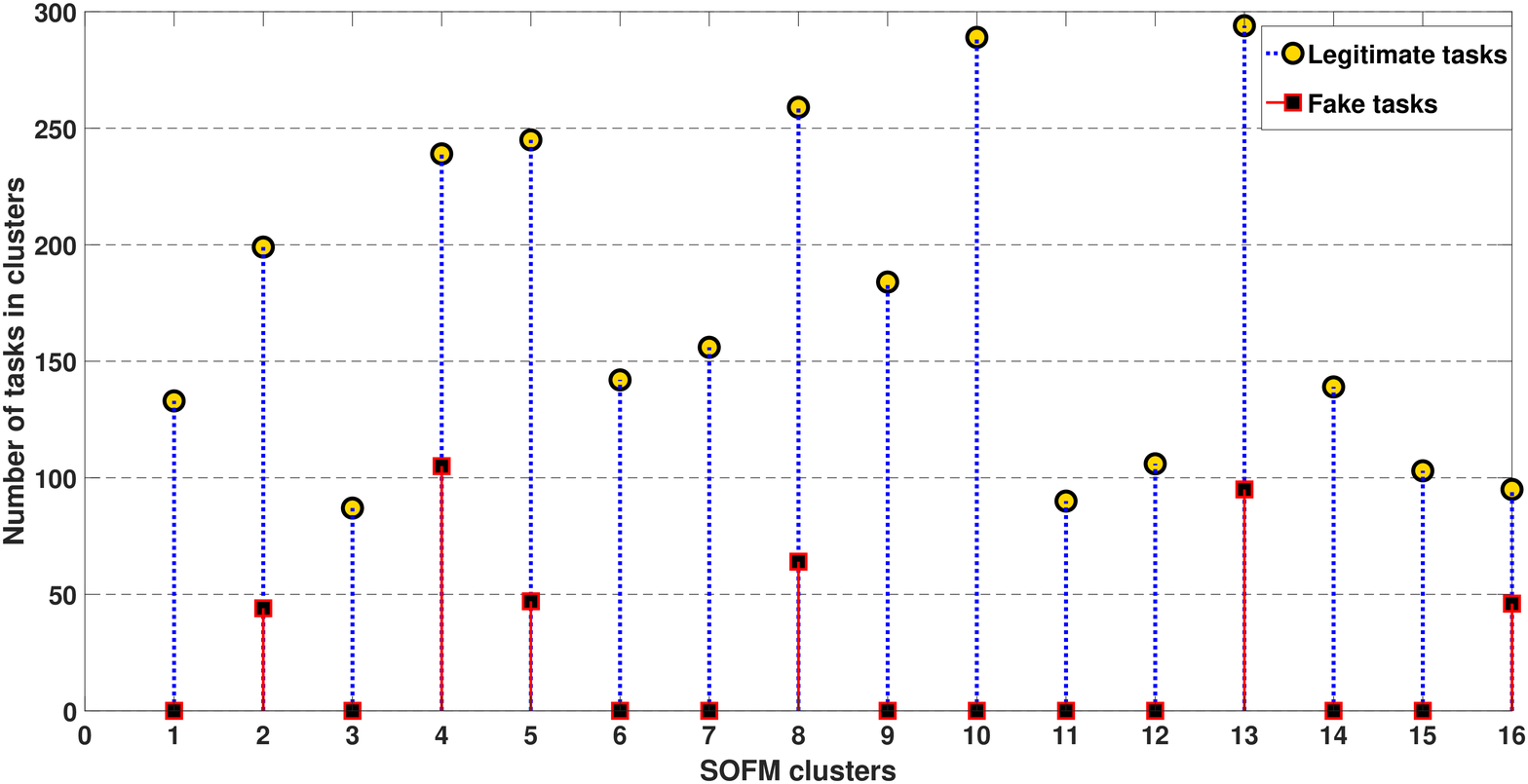}
        \caption{Test}
        \label{fig:sofm_clusters_test}
    \end{subfigure}
        \caption{SOFM cluters for 16 neurons (4$\times$4) to identify legitimate-only (1,3,6,7,9,10,11,12,14,15)  and mixed (legitimate \& fake) clusters in training and test phase. }
	\label{fig:sofm_clusters}
\end{figure}

\subsection{Self Organizing Feature  Map (SOFM) for pre-clustered Data}
SOFM is an artificial neural network-based model that is trained in an unsupervised manner; thus, multiple clusters of similar data can be covered by a pre-defined number of neurons. Neurons in a SOFM are organized as 2D and they are updated through weights which are equal to the number of inputs. When a sample is applied to SOFM, inputs of each sample and weights of each neuron are compared to each other through a distance-based metric. The winner neuron which is the closest neuron to the inputs is called as Best Matching Unit (BMU). The weights of BMU are updated to the closer to inputs of sample data. Therefore, SOFM organizes itself according to the whole samples in the dataset. Algorithmic details about SOFM can be seen in \cite{Zhang2021Empowering,Simsek2020Self}.

As seen in Fig. \ref{fig:algorithm}, SOFM generates two sets of clusters: legitimate-only and mixed. The latter contains legitimate and fake tasks together for the training dataset. As seen in Fig. \ref{fig:sofm_clusters}, similar clustering results are obtained by training and testing data. Mixed dataset that contains legitimate and fake tasks together is utilized in the pre-clustered Deep Neural Network (PrecDeepNN) to yield a more specific prediction output on the test set. Although fake tasks are always outnumbered by the legitimate tasks in the  MCS  dataset (which is a realistic assumption), less imbalanced data can be obtained from mixed data through extracting pre-clustered legitimate-only data out of the training data.

SOFM can generate legitimate-only data through clusters $\{1,3,6,7,9,10,11,12,14,15\}$ so legitimate classification is realized by pre-clustered legitimate tasks process as shown in Fig. \ref{fig:algorithm}. In addition to clustering legitimate-only data, SOFM can generate mixed data through clusters $\{2,3,4,8,13,16\}$ to train the pre-clustered DeepNN as depicted in  Fig. \ref{fig:algorithm}. The mixed data consists of legitimate and fake tasks; however, legitimate data rates are mitigated in this new training dataset so fake tasks can be classified better than the original training dataset. In the test phase, SOFM generates not only legitimate-only tasks which are used for classification performance of the test data but also mixed tasks for prediction performance of PrecDeepNN in Fig. \ref{fig:algorithm}.

\begin{figure*}[!t]
        \centering
        \includegraphics[width = 1.0\textwidth]{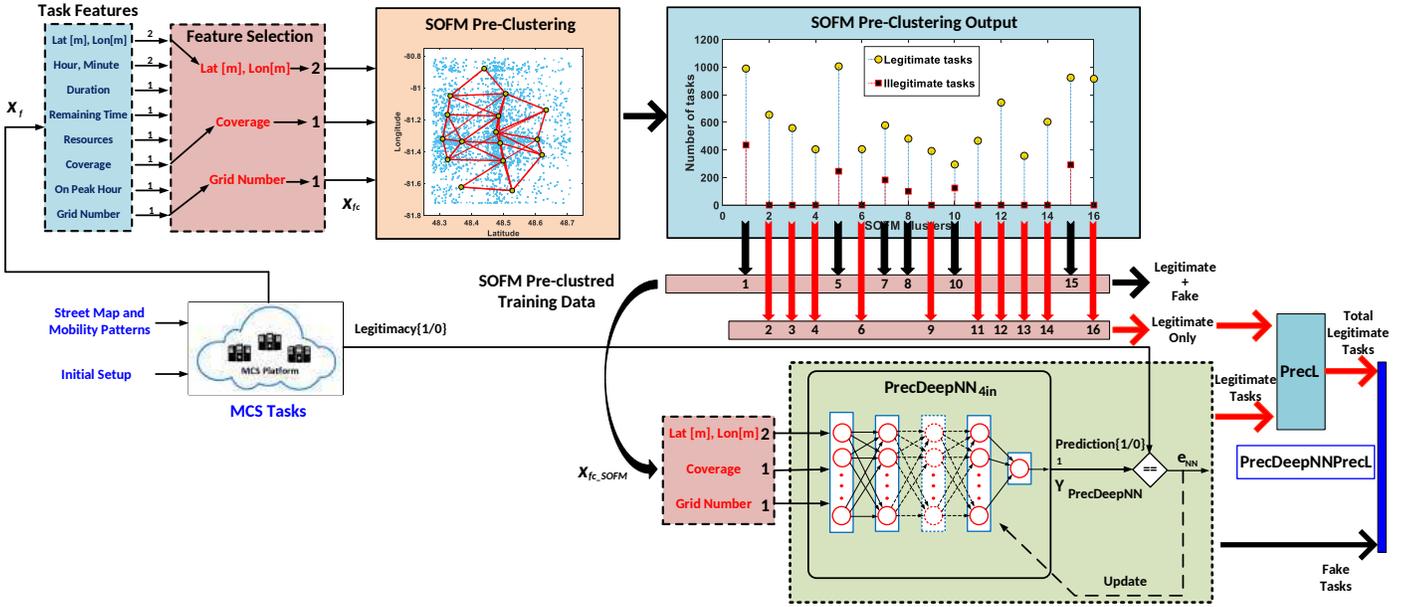}
        \caption{Pre-clustering based DeepNN process for training and test. Cluster numbers are just for illustrative purposes.}
        \label{fig:deep_NN_preclustering}
\end{figure*}

\subsection{Pre-clustered Deep Neural Network (PrecDeepNN) Model}

The data flow over a PrecDeepNN and its interaction with SOFM-driven pre-clustered data can be seen in Fig. \ref{fig:algorithm}. The aim of this model is to be trained by pre-clustered training data whose legitimate tasks are mitigated by SOFM driven pre-clustering. After the training is completed, testing data that are pre-clustered as mixed data are used as the input for PrecDeepNN. The final prediction performance of the model is combined with legitimate-only task clusters output by SOFM as depicted in Fig. \ref{fig:deep_NN_preclustering}.

Training of the PrecDeepNN model is illustrated in (\ref{eq:precdeepnn_training1}) and error definition of the PrecDeepNN model is depicted in (\ref{eq:precdeepnn_training2}) in which $x_{fc\_SOFM}$ and $Y_{PrecDeepNN}$ are the pre-clustered inputs by SOFM and output of the PrecDeepNN model, respectively. $L_{MCS}(x_{fc\_SOFM})$ denotes legitimacy of $x_{fc\_SOFM}$
in (\ref{eq:precdeepnn_training2}).   

\begin{equation}
 \label{eq:precdeepnn_training1}
 \begin{split}
\left(PrecDeepNN\right)_{T}= \arg \mathop {\min }\limits_i \left\|{\begin{array}{*{10}{c}} \cdots &{{e_{NN}^{(i)}}^T}& \cdots \end{array}}\right\|   \\
i = 1, 2, ..., N.
\end{split}
\end{equation}

\begin{equation}
\label{eq:precdeepnn_training2}
{e_{NN}^{(i)}} = L_{MCS}\left(x_{fc\_SOFM}^{(i)}\right) - Y_{PrecDeepNN}\left(x_{fc\_SOFM}^{(i)}\right)
\end{equation}

Output of the PrecDeepNN model after the training process and given the pre-clustered inputs by SOFM is defined in (\ref{eq:precdeepnn_training3}) where ${PrecDeepNN}_{T}$ indicates trained PrecDeepNN model.

\begin{equation}
\label{eq:precdeepnn_training3}
Y_{PrecDeepNN}={PrecDeepNN}_{T}\left(x_{fc\_SOFM}\right)
\end{equation}

Legitimate and fake task prediction results of proposed PrecDeepNNPrecL model are depicted in (\ref{eq:precdeepnn_training4}) and (\ref{eq:precdeepnn_training5}), respectively. $P_L\{{PrecDeepNN}_{T}\}$ indicates legitimate task prediction of the PrecDeepNN model and $L_{Prec}$ indicates the pre-clustered legitimate test data by SOFM in (\ref{eq:precdeepnn_training4}). $P_F\{{PrecDeepNN}_{T}\}$ denotes fake task prediction of the PrecDeepNN model in (\ref{eq:precdeepnn_training5}).

\begin{equation}
\label{eq:precdeepnn_training4}
\resizebox{0.9\hsize}{!}{$Leg\_tasks$= $P_L\{{PrecDeepNN}_{T}\left(x_{fc\_SOFM}\right)\} + L_{Prec}$}
\end{equation}

\begin{equation}
\label{eq:precdeepnn_training5}
Fake\_tasks= P_F\{{PrecDeepNN}_{T}\left(x_{fc\_SOFM}\right)\}
\end{equation}

The general structure of the proposed method and input-output relationships can be seen in Fig. \ref{fig:deep_NN_preclustering}.

\section{Performance Evaluation}
\label{sec:numresults}

In this study, Timmins region in Canada is chosen to collect tasks generation data from MCS platform.  Crowdsensim that is a mobile crowd sensing simulator generates 10 features for submitted tasks.  While the centers of fake task submission zones with 200m radius are selected randomly, fake/legitimate task proportion is set to 1/9 for generating 4,000 sensing tasks. The total number of data is 14306 in the six-days campaign and the first 80\% of the total data is used for training dataset (11,145) and remaining 20\% of the total data is selected for test dataset. Since test dataset is not selected inside the training dataset, more challenging extrapolation performance can be obtained from test dataset to demonstrate the efficiency of the proposed methodology.

Simulation results have been obtained by MATLAB code generating environment and  the neural  network  Toolbox. latitude,  longitude,  coverage  and  grid  number  features  has been   selected  through ReliefF sequential feature selection algorithm is used in this study by folloing the same setting in our previous work \cite{Simsek2020Self}. In addition, DeepNN is modelled with four hidden layers where each hidden layer has 15 neurons that was  \cite{Simsek2020Self}. Since training and test sets  are totally  different  than  each  other, all  numerical results have been collected as extrapolation performance.

\begin{table*}[!hbt]
\caption{SOFM Clustering Results of Training and Testing Datasets for pre-clustered DeepNN with pre-cluster Legitimate Tasks}
\label{tab:sofm_clusters}
\resizebox{\textwidth}{!}{%
\begin{tabular}{c|c|ccccccccccccccccc}
\hline
Dataset                                                                   & Tasks                                                     & \multicolumn{16}{c}{SOFM Clusters}                                                                                                                                                                                                                                                                                                                                                 & pre-clustered                                                                      \\
                                                                          &                                                           & 1   & 2                                                 & 3   & 4                                                 & 5                                                 & 6   & 7   & 8                                                 & 9   & 10   & 11  & 12  & 13                                                & 14  & 15  & 16                                                &                                                                                    \\ \hline
\multirow{2}{*}{\begin{tabular}[c]{@{}c@{}}Training \\ Data\end{tabular}} & Legitimate-only                                           & 418 &                                                   & 320 &                                                   &                                                   & 466 & 653 &                                                   & 702 & 1056 & 203 & 237 &                                                   & 603 & 559 &                                                   & \begin{tabular}[c]{@{}c@{}}PrecL\\ 5217\end{tabular}                               \\ \cline{2-19} 
                                                                          & \begin{tabular}[c]{@{}c@{}}Legitimate\\ Fake\end{tabular} &     & \begin{tabular}[c]{@{}c@{}}640\\ 100\end{tabular} &     & \begin{tabular}[c]{@{}c@{}}989\\ 435\end{tabular} & \begin{tabular}[c]{@{}c@{}}863\\ 182\end{tabular} &     &     & \begin{tabular}[c]{@{}c@{}}841\\ 245\end{tabular} &     &      &     &     & \begin{tabular}[c]{@{}c@{}}923\\ 292\end{tabular} &     &     & \begin{tabular}[c]{@{}c@{}}294\\ 124\end{tabular} & \begin{tabular}[c]{@{}c@{}}PrecDeepNN\\ Training set\\ 4550 \\ 1378\end{tabular} \\ \hline
\multirow{2}{*}{\begin{tabular}[c]{@{}c@{}}Test \\ Data\end{tabular}}     & Legitimate-only                                           & 133 &                                                   & 87  &                                                   &                                                   & 142 & 156 &                                                   & 184 & 289  & 90  & 106 &                                                   & 139 & 103 &                                                   & \begin{tabular}[c]{@{}c@{}}PrecL\\ 1429\end{tabular}                               \\ \cline{2-19} 
                                                                          & \begin{tabular}[c]{@{}c@{}}Legitimate\\ Fake\end{tabular} &     & \begin{tabular}[c]{@{}c@{}}199\\ 44\end{tabular}  &     & \begin{tabular}[c]{@{}c@{}}239\\ 105\end{tabular} & \begin{tabular}[c]{@{}c@{}}245\\ 47\end{tabular}  &     &     & \begin{tabular}[c]{@{}c@{}}259\\ 64\end{tabular}  &     &      &     &     & \begin{tabular}[c]{@{}c@{}}294\\ 95\end{tabular}  &     &     & \begin{tabular}[c]{@{}c@{}}95\\ 46\end{tabular}   & \begin{tabular}[c]{@{}c@{}}PrecDeepNN\\ Test set\\ 1331\\ 401\end{tabular}       \\ \hline
\end{tabular}%
}
\end{table*}

SOFM clustering results show that 5,217 pre-clustered legitimate-only data can be separated from total training data. Remaining 4,550 legitimate and 1,378 fake tasks data constitute 5,928 mixed data to train the PrecDeepNN model. In the test phase, 1,428 PrecL data points which are pre-clustered as legitimate-only before test phase are appended to the prediction output of the PrecDeepNN, thus proposed PrecDeepNNPrecL model testing performance can be obtained from the whole process. Before testing performance of PrecDeepNN, 1,331 legitimate and 401 fake tasks are evaluated by the model to generate the prediction results. All numerical values on pre-clustering performance of SOFM are presented in Table \ref{tab:sofm_clusters}.  

\begin{figure}[!t]
        \centering
        \includegraphics[width = 0.5\textwidth, trim=2cm 0.0cm 2cm 0.0cm, clip]{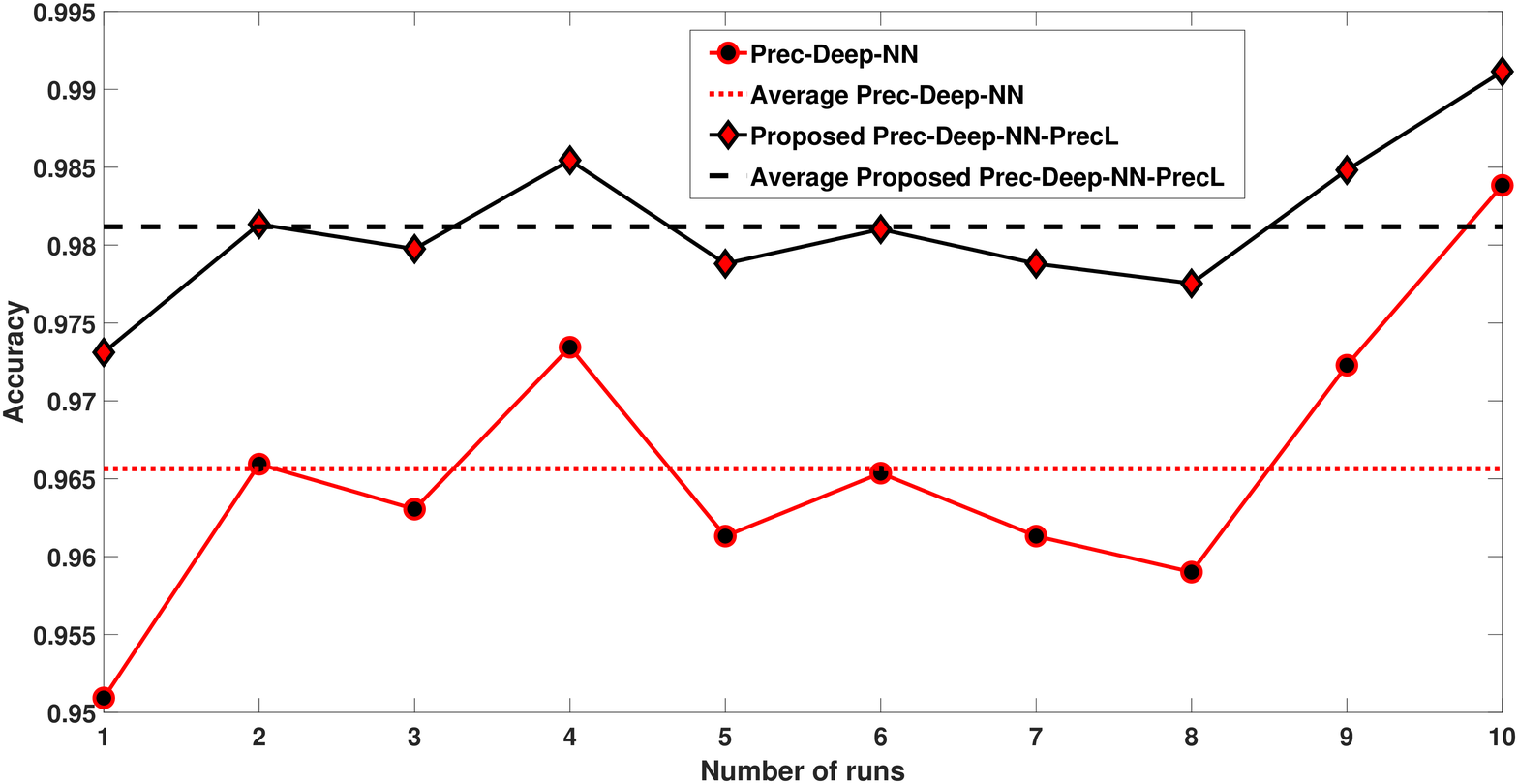}
        \caption{In the proposed methodology, pre-clustered Deep-NN results as average accuracy of 0.9656 is improved to 0.9812 as average accuracy of pre-clusterd Deep-NN with pre-clusterd legitimate tasks through adding pre-clustered legitimate-only tasks which are obtained by SOFM clusters 1,3,6,7,9,10,11,12,14,15 in Table \ref{tab:sofm_clusters}. }
        \label{fig:proposed_results}
\end{figure}

\begin{figure}[!t]
        \centering
        \includegraphics[width = 0.5\textwidth,trim=2cm 0.0cm 2cm 0.0cm, clip]{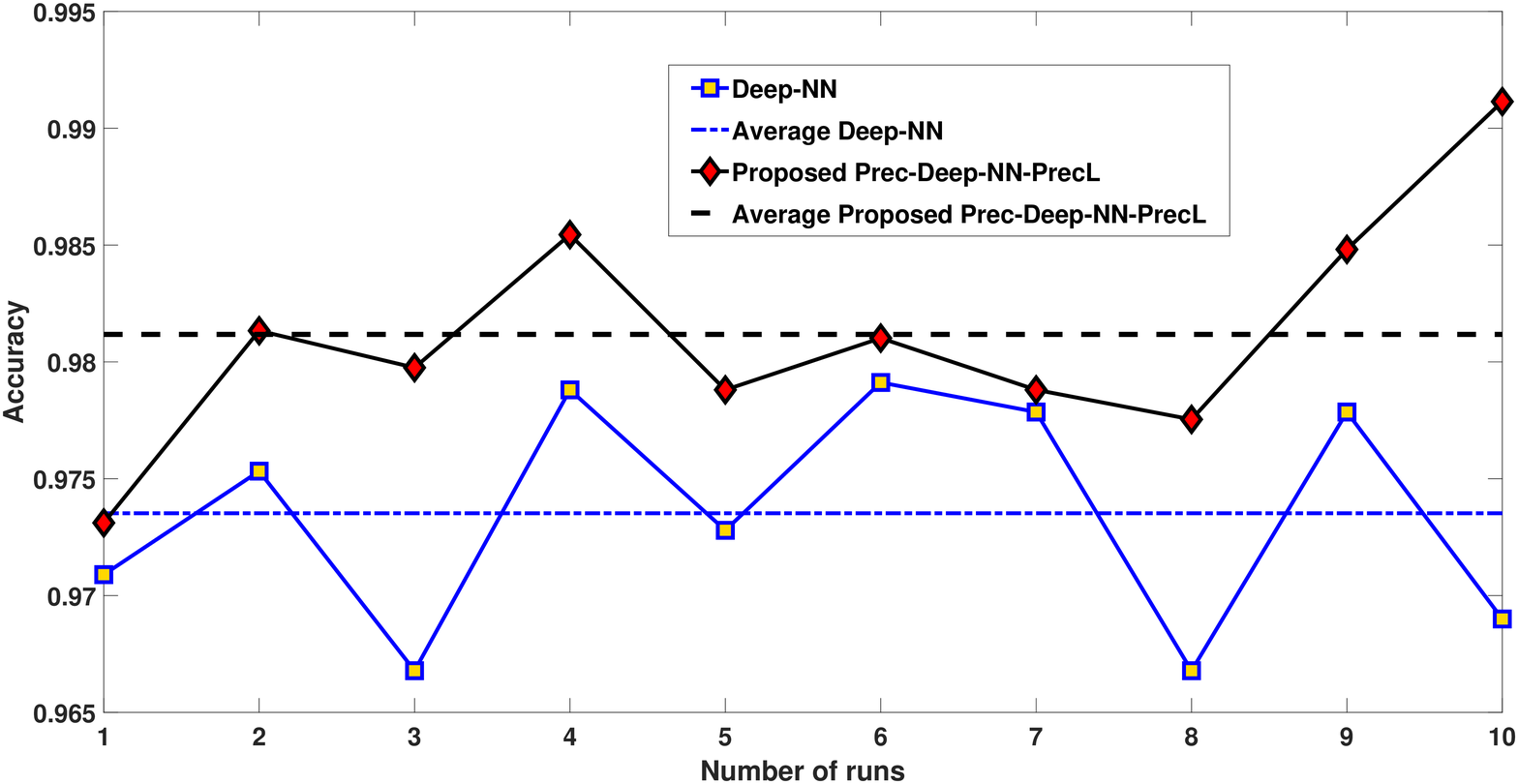}
        \caption{Pre-clustered legitimate tasks are appended to the prediction performance of the pre-clustered Deep-NN. The integrated model  results in an average accuracy of 0.9812 whereas the classical Deep-NN results in 0.9735 accuracy.}
        \label{fig:result_comparison}
\end{figure}

The test performance of PrecDeepNN and PrecDeepNNPrecL are compared in Fig. \ref{fig:proposed_results} to demonstrate the contribution of PrecL that is appended to PrecDeepNN to obtain the proposed PrecDeepNNPrecL results. Overall testing improvement between the initial accuracy of DeepNN and final accuracy of PrecDeepNNPrecL is demonstrated in Fig. \ref{fig:result_comparison}. This figure verifies the efficiency of the proposed methodology since 0.9735 average accuracy is increased to 0.9812 after applying the PrecDeepNNPrecL model.

\section{Conclusion}
\label{sec:conclusion}
Task submission and data collection in MCS have led to vulnerabilities against adversaries. Sensing resources of participant devices and MCS platforms can be drained by fake task submissions. In this paper, the supervised DeepNN model is combined with an unsupervised SOFM method to improve the task legitimacy detection performance of the MCS system. Pre-clustered legitimate-only tasks are initially clustered by SOFM, and the remaining (i.e., mixed) data are used to train the DeepNN model. Test results of obtained from the PrecDeepNN model can predict fake and legitimate tasks with an average accuracy of 0.9656. Pre-clustered legitimate-only test data that are initially clustered by SOFM are appended to the prediction results of the PrecDeepNN so PrecDeepNNPrecL has achieved an overall average accuracy of 0.9812. Hence, the initial average accuracy of DeepNN (0.9735) is improved up to an average of 0.9812 by the proposed PrecDeepNNPrecL. 

The improvement of the proposed methodology is sufficient to demonstrate the efficiency of PrecDeepNNPrecL, however our ongoing agenda includes studying the detection performance impact of different structures as well as various distributions of legitimate and fake tasks in the MCS platform. In addition to these, our ongoing agenda includes implementation of the proposed defense mechanism on a realistic test bed under a real application such as vehicular crowdsensing. Furthermore, collaboration between fake task submitters have not been considered in our threat model, which remains an open issue that is currently included in our ongoing research agenda. Last but not least, incorporation of a game theoretic model between the MCS platform and the task submitters can help eliminate the submission of fake tasks even before the detection module.  Therefore, building on the existing literature, we are planning to bridge a game theoretic incentive-based approach and the proposed fake task detection scheme in the extension of this study.
 \section*{Acknowledgment}
 This work was supported in part by the the Natural Sciences and Engineering Research Council of Canada (NSERC) under the DISCOVERY Program, Canada Research Chairs Program and NSERC CREATE TRANSIT Project. 

\bibliographystyle{ieeetr}

\end{document}